\documentclass[conference]{IEEEtran}

\usepackage[utf8]{inputenc}

\usepackage{amssymb}
\usepackage[normalem]{ulem}
\usepackage[table]{xcolor}

\usepackage{balance}

\IEEEoverridecommandlockouts 
\usepackage{graphicx}
\usepackage{algorithmic}
\usepackage{algorithm}
\usepackage{listings}
\usepackage[OT4,T1]{fontenc}
\usepackage[cmex10]{amsmath}
\usepackage{url}

\usepackage{multirow}

\title{Higher-Order Quantum-Inspired Genetic Algorithms\thanks{This work was supported in part by PL-Grid Infrastructure}}
\author{
\IEEEauthorblockN{Robert Nowotniak, Jacek Kucharski}
\IEEEauthorblockA{Institute of Applied Computer Science\\
Lodz University of Technology\\
18/22 Stefanowskiego St., 90-924 Lodz, Poland\\
Email: \{rnowotniak,jkuchars\}@kis.p.lodz.pl}
}

\definecolor{brick}{rgb}{0.7,0.7,0.7}

\begin{document}
\maketitle              

\begin{abstract}
This paper presents a theory and an empirical evaluation of Higher-Order Quantum-Inspired Genetic Algorithms.
Fundamental notions of the theory have been introduced, and a~novel Order-2 Quantum-Inspired
Genetic Algorithm (QIGA2) has been developed.
Contrary to all QIGA algorithms which represent quantum genes as independent qubits,
in higher-order QIGAs quantum registers are used to represent genes strings,
which allows modelling of genes relations using quantum phenomena.
Performance comparison has been conducted on a benchmark of 20 deceptive combinatorial optimization problems.
It has been presented that using higher quantum orders is beneficial for genetic algorithm
efficiency, and the new QIGA2 algorithm outperforms the old QIGA algorithm tuned in highly
compute-intensive metaoptimization process.

\end{abstract}

\section{Introdution}

\IEEEPARstart{R}{esearch}
on quantum-inspired computational intelligence techniques was started
by Narayann[1] in 1996, and the first proposal of Quantum-Inspired Genetic Algorithm (QIGA1)
has been presented by Han and Kim in [2].
Quantum-Inspired Genetic Algorithms belong to a new class of
artificial intelligence techniques, drawing inspiration from both
evolutionary[3] and quantum[4] computing. 
Current literature on the subject consists of about a few hundreds scientific papers.
Only a few papers attempt to theoretically analyse the properties of that class of algorithms.
Among those there are i.a. [22,28],
which has been emphasized in conclusions of recent comprehensive
surveys [18,29].

In QIGA algorithms,
representation and genetic operators are based on 
computationally useful aspects of both
biological evolution and unitary evolution of quantum systems. QIGA algorithms
use quantum mechanics concepts including qubits and superposition of
states.
QIGA algorithms have been successfully applied to a broad range of search and
optimization problems[5,6,7].
The algorithms have demonstrated their particular efficacy for solving complex
optimization problems. Recent years have witnessed successful applications of
Quantum-Inspired Genetic Algorithms in a variety of fields,
including image processing[8,9,10], flow shop scheduling[11,12], thermal unit
commitment[13,14], power system optimization[15,16], localization of mobile
robots[17] and many others.

For a current and comprehensive survey of Quantum-Inspired Genetic Algorithms
and the necessary background of Quantum Computing and Quantum-Inspired
Computational Intelligence techniques, the reader is referred to [1,2,18,29].


This paper is structured as follows.
In Section 1, an introductory background and the most important references for the subject field have been given.
In Section 2, the theory of Higher-Order Quantum-Inspired Genetic Algorithms has been presented.
In Section 3, details of the original Order-2 Quantum-Inspired Genetic Algorithm have been provided.
In Section 4, experimental results have been provided and evaluated.
In Section 5, the article has been briefly summarized,
final conclusions have been drawn, and also possible directions for future research have been suggested.

\newtheorem{definition}{Definition}

%
%
%

\section{Theory of Higher-Order Quantum-Inspired Genetic Algorithms}

Let $N\in\mathbb{N}^+$ denote the length of chromosomes in the algorithm (i.e. problem size),
$X$ -- search space of the optimization problem, $Q$ -- quantum population
(a set of quantum individuals in QIGA algorithm), and $P$ -- classical population (a set of elements in $X$ space).
Let us assume that each individual in the algorithm consists of a single quantum chromosome.\\
We introduce the following new notions.

\begin{definition}[quantum order $r\in\mathbb{N^+}$] the size of the biggest quantum register
used in the algorithm.
\begin{equation}
	1 \le r \le N
\end{equation}
\end{definition}
We say an algorithm is Order-$r$, if $r$ is the size of the biggest
quantum register used in that algorithm.
\uline{All Quantum-Inspired Genetic Algorithms that use independent qubits
to represent binary genes are Order-1.}
All existing algorithms, presented in the literature so far
are Order-1 in terms of this theory.
To simplify the further discussion, let us assume that all quantum registers
used in the algorithm have the same size.

\begin{definition}[relative quantum order $w$] -- the ratio of quantum order
$r$ to quantum chromosomes length $N$ (problem size) in the algorithm.
\begin{equation}
	w = \frac{r}{N} \in (0,1]
\end{equation}
\end{definition}

If a certain QIGA algorithm uses a representation of solutions based on 100 independent qubits
(binary quantum genes), the relative quantum order for that algorithm is
$w = \frac{1}{100}$.
If the size of a problem (the number of binary variables)
is $N = 60$, and the representation is based on 3-qubit registers, then the relative quantum range
is $w = \frac{3}{60} = 0.05$ etc.

The algorithms characterised by $w=1$ are "true" quantum algorithms,
where a single quantum register contains all the binary variables.
For $w=1$, when the number of binary variables (the size of the problem $N$) grows linearly,
the cost of simulation grows exponentially (which corresponds to a simulation of a real quantum computer). 

\begin{figure}[tbp]
	\centering
	\vspace*{5mm}
	\includegraphics[width=\hsize]{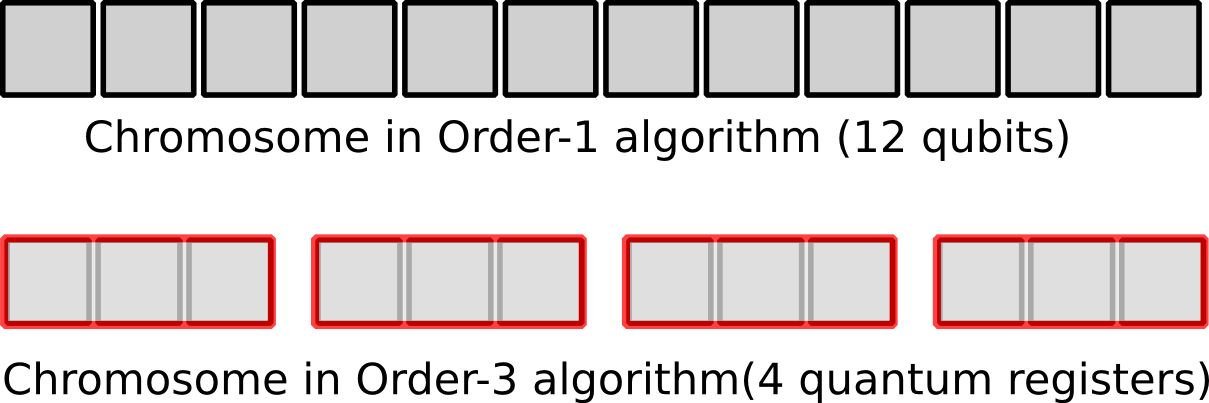}
	\caption{
	Examples of chromosomes of length $N=12$ for Order-1 and Order-3 algorithms.
	Consecutive genes are joined to $r$-qubit quantum registers.
	In Order-1 algorithm, the chromosome consists of 12 independent qubits,
	each one is a unit vector in 2-dimensional space.
	In Order-3 algorithm, the chromosome consists of 4 quantum registers,
	each one is a unit vector in $2^3 = 8$-dimensional space.
	\label{fig:qgenes-groups}}
\end{figure}

\begin{definition}[quantum factor $\lambda\in\lbrack 0,1 \rbrack$]

For a given algorithm, the quantum factor is defined as a ratio of the dimension of space in a given
class of algorithms to the dimension of space of the full quantum register of $N$ qubits. 
Additionally, if there are no quantum elements in the algorithm (e.g. a simple genetic algorithm SGA[30],
operating in a discrete space of binary strings), then  $\lambda= 0$.\\
Thus, the numerical value of the factor is expressed as:
\begin{equation}
\lambda = \frac{2^r \frac{N}{r}}{2^N} = \frac{2^r}{w 2^N}
\end{equation}
where $r$ is the quantum order of an algorithm and $N$ is the problem size.
The $2^r$ in the numerator of the above formula corresponds to the dimension of the state space in the $r$-qubit quantum register (the biggest quantum register used in an algorithm of that class). Such quantum register codes a $2^r$-point probability distribution (it shows the probability of choosing one from $2^r$ elements of a solution space $X$). 
$2^N$ corresponds to the dimension of the state space of a quantum register containing all $N$ qubits.
\end{definition}

\begin{figure}[h]
\includegraphics[width=\hsize]{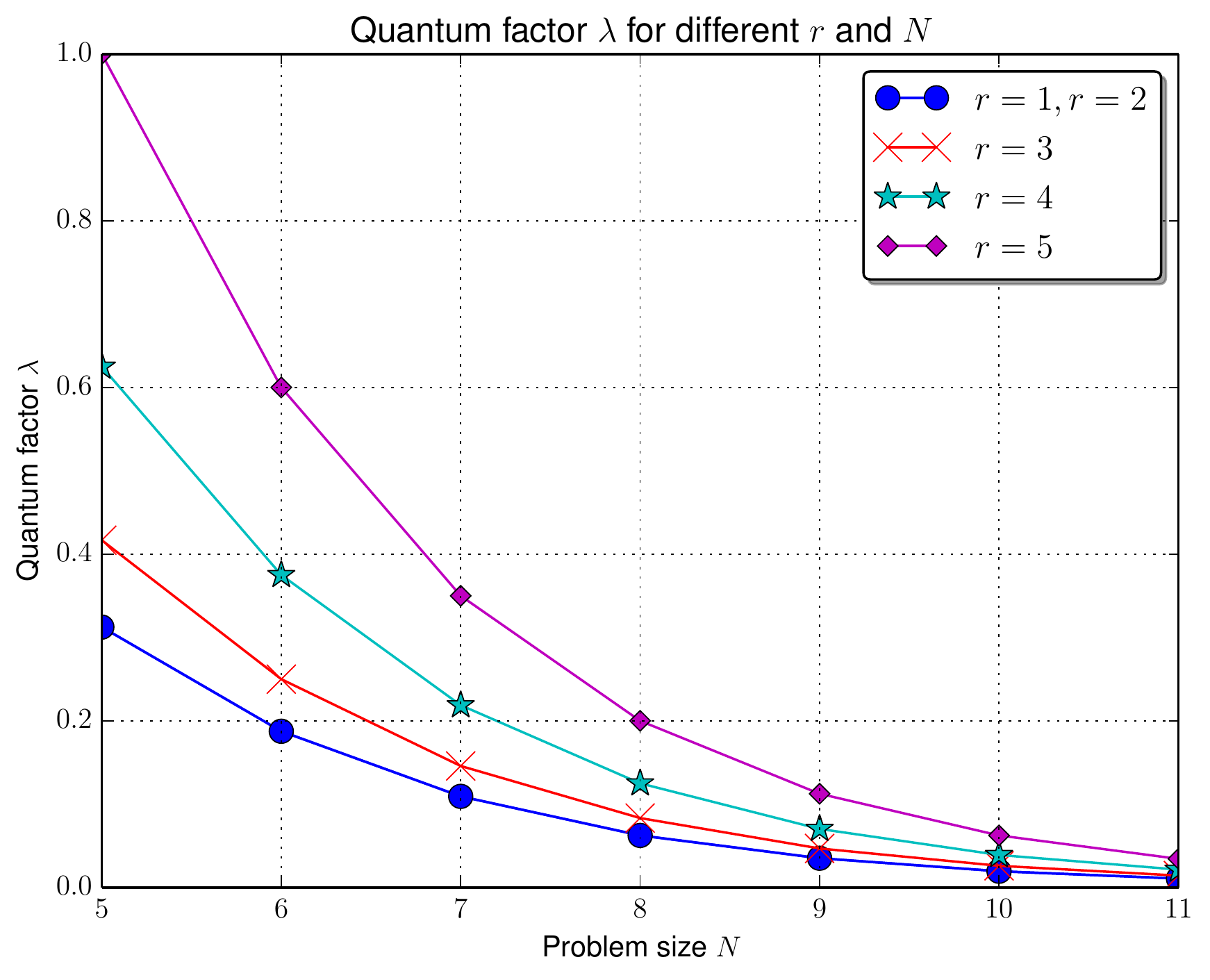}
\caption{
Quantum factor $\lambda$ for different problem size $N$ and different quantum orders
$r\in\{1,2,3,4,5\}$\label{fig:orders}}
\end{figure}

In Order-1 algorithms, chromosomes consist of $N$ independent qubits.
According to the Quantum Computing theory the state of each qubit is described by a unit vector in a 2-dimensional space
($|q\rangle=[\alpha~~\beta]^T$),
so the space dimension for the chromosomes in such algorithms is $2^r \frac{N}{r}=2N$.

In Order-2 algorithms, chromosomes consist of $\frac{N}{2}$ size-2 quantum registers.
The state for each register is described by a unit vector in a 4-dimensional space
($|q\rangle=[\alpha_0~\alpha_1~\alpha_2~\alpha_3]^T$).
Therefore, the dimension of space for the chromosomes in such algorithms is also $2^2 \frac{N}{2}=2N$.
However, in Order-1 algorithms only one qubit coordinate might be independently modified (one degree of freedom), while in Order-2 algorithms the same can be done with 3 out of 4 coordinates of the 2-qubit quantum register state. 
Consequently, it allows for modelling of relations between two neighbouring genes joined in a common register. 

\textbf{For even Higher-Order algorithms ($r\ge 3$), simulating quantum element makes the algorithm
exponential computational complexity.}
Relationship between quantum factor $\lambda$, quantum order $r$ for growing problem size $N$
has been presented in Figure~\ref{fig:orders}.

It should be noted that for $r=1$ (all regular Order-1 QIGA algorithms):
\[
\lambda = \frac{2^1 \frac{N}{1}}{2^N} = \frac{2 \cdot N}{2^N}
\]
Thus, for example, in an algorithm coding solutions in the form of 10-element strings of independent qubits,
$\lambda = \frac{20}{2^{10}} \approx 0.02$.
It means that the size of space in such algorithm comprises ~2\% of the full quantum register state space, which would include 10 binary variables. Together with the increase of size of a problem $N$ and for a constant quantum order $r=1$, the quantum factor decreases exponentially and becomes  $\lambda < 10^{-10}$ for $N=50$.


For that reason, for a constant quantum order $r=1$ (QIGA Order-I quantum-inspired algorithms)
and for an increasing size of a problem $N$, the quantum factor $\lambda$ has a limit that equals zero:
\[
\lim_{\substack{r=1\\ N\rightarrow \infty}} \lambda = \lim_{\substack{r=1\\ N\rightarrow \infty}} \frac{2\cdot{}N}{2^N} = 0
\]
However, for $r=N$ (typical quantum algorithms)
\begin{equation}
\lambda = \frac{2^N \frac{N}{N}}{2^N} = \frac{2^N}{2^N} = 1
\end{equation}
For $\lambda= 1$, when the number of variables (the size of a problem $N$) grows linearly,
the cost of simulation grows exponentially (which corresponds to a full simulation of a real quantum computer).\\
Thus, algorithms can be classified according to quantum factor $\lambda$ value as follows:
\begin{enumerate}
\item $\lambda = 0$ -- a classical algorithm without any quantum elements,
operating in a discrete finite space (e.x. SGA[30] operating in finite
discrete binary strings space).
\item $\lambda \in (0,1)$ -- a quantum-inspired algorithm, like QIGA1 (order $r=1$),
or higher-order algorithm.
\item $\lambda = 1$ -- a "true" quantum algorithm which requires either a real
quantum level hardware, or an exponential complexity simulation on classical computer.
\end{enumerate}

\textbf{Order-$r$ Quantum-Inspired Genetic Algorithms are capable of modelling relations between
separate genes which are joined into the same quantum register of size $r$.
This allows the algorithm to work better for deceptive combinatorial optimization
problems and to better solve strong epistasis in deceptive problems.}
This is presented empirically the next sections of the paper.

%
%
%
%

\section{Order-2 Quantum-Inspired Genetic Algorithm}

In this section, a novel Order-2 Quantum-Inspired Genetic Algorithm (QIGA2) has been
presented. The algorithm has been developed based on the theory of higher-order
quantum-inspired algorithms presented in the previous section.

Pseudocode of the algorithm has been presented in Algorithm 1,
and in general it is very similar to a typical evolutionary algorithm scheme.
The general principle of operation of the algorithm is very similar to the initial QIGA 1 algorithm, but instead of independent qubits modelling successive binary genes, the QIGA 2 algorithm uses 2-qubit quantum registers representing successive pairs of genes.

In each generation of the algorithm a classic population $P$ (a set of elements from the solution space $X$) is sampled
through observation of quantum states of the quantum population $Q$ i.e. $|P|$-times repeated sampling of the space $X$ according to probability distributions stored in $Q$.
The classical population $P$ is then evaluated exactly as in a typical evolutionary algorithm.
The quantum population $Q$, however, is updated in consecutive generations in such a way that it increases the probability of sampling the best solution $b$ neighbourhood,
which has been recorded in previous generations of $P$.


\begin{algorithm}[tbp]
\caption{Order-2 Quantum-Inspired Genetic Algorithm}
\begin{algorithmic}[1]

\STATE $t \leftarrow 0$\;
\STATE Initialize quantum population $Q(0)$\;
\WHILE{$t \le t_{max}$}
\STATE $t \gets t+1$\;
\STATE Generate $P(t)$ by observing quantum pop. $Q(t-1)$\;
\STATE Evaluate classical population $P(t)$\;
\STATE Update $Q(t)$\;
\STATE Save best classical individual to $b$\;
\ENDWHILE
\end{algorithmic}
\end{algorithm}

The key new elements distinguishing QIGA2 from the previous Order-1 algorithms are the modified method of representing solutions
and the new genetic operators working in a space of a higher dimension and described by $4\times 4$ unitary matrices in the quantum-mechanic sense.
Both original elements have been described in the next subsections respectively.

\subsection{Representation of solutions in QIGA2}

The fundamental difference between the already existing QIGA1 and QIGA2 algorithms lies in the way they represent solutions.
In QIGA1 algorithms, quantum genes are modelled with qubits i.e. two-level quantum systems
$|q\rangle=\alpha|0\rangle + \beta|1\rangle=[\alpha~~\beta]^T$
which are able to code two-point probability distributions.
It corresponds to a possibility of each gene to have a value 0 or 1 with a probability of $|\alpha|^2$ and $|\beta|^2$ accordingly.
It has been depicted in Figure 3.

\uline{In the authors' QIGA2 algorithm, the representation of solutions is based on using the adjacent 2-qubit quantum registers.}
For that purpose the adjacent genes are consecutively paired.
The corresponding 2-qubit registers $|q\rangle=[\alpha_0~~\alpha_1~~\alpha_2~~\alpha_3]^T$ code 4-point probability distributions.
So, in a single quantum register 4 values of probability $|\alpha_0|^2$, $|\alpha_1|^2$, $|\alpha_2|^2$, $|\alpha_3|^2$ are recorded.
These are probabilities of having a value of 00, 01, 10 and 11 for each given pair of genes accordingly. It is presented in Figure 4.
Similarly to QIGA1 algorithms, the proposed QIGA2 uses only the real parts of probability amplitudes. It ignores the imaginary part of amplitudes $\alpha_0, \ldots, \alpha_3$.

At the stage of the $Q(0)$ base population initialization, all genes can be given the value of $q_{ij} = [\frac{1}{2}~~\frac{1}{2}~~\frac{1}{2}~~\frac{1}{2}]^T$,
which corresponds to a situation when the algorithm samples the entire solution space $X$ with the same probability.    



\begin{figure}[tbp]
\includegraphics[width=\hsize]{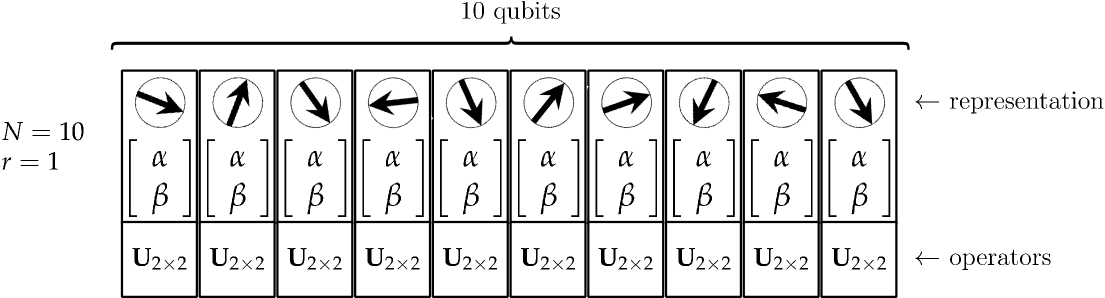}
\caption{In QIGA1, representation is based on isolated qubits / binary quantum genes\label{fig:repr_qiga1}}
\end{figure}

\begin{figure}[tbp]
\includegraphics[width=\hsize]{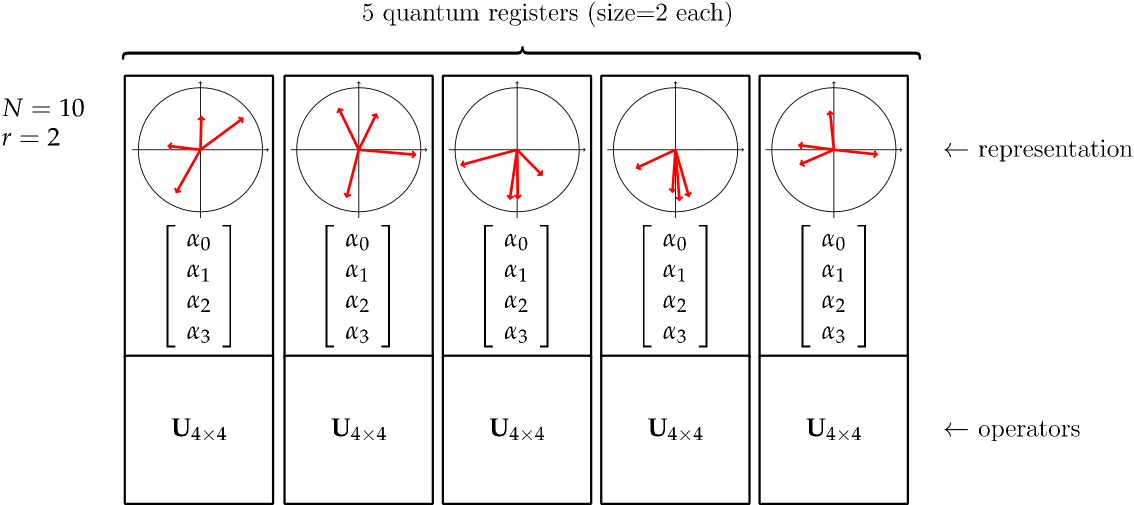}
\caption{In QIGA2, quantum registers are used to represent pairs of genes\label{fig:repr_qiga2}}
\end{figure}

\begin{algorithm}[tbp]
\caption{Observation of genes pair in QIGA2}
\begin{algorithmic}[1]
\REQUIRE $q_{ij} = [\alpha_0~~\alpha_1~~\alpha_2~~\alpha_3]^T$ -- quantum register of 2 qubits

\STATE $r \gets \textrm{uniformly random number from [0,1]}$\;
\IF{$r < |\alpha_0|^2$ }
\STATE $p \gets 00$\;
\ELSIF{$r < |\alpha_0|^2 + |\alpha_1|^2$}
		\STATE $p \gets 01$\;
\ELSIF{$r < |\alpha_0|^2 + |\alpha_1|^2 + |\alpha_2|^2$}
		\STATE $p \gets 10$\;
\ELSE
		\STATE $p \gets 11$\;
\ENDIF

\end{algorithmic}
\end{algorithm}

\subsection{Order-2 quantum genetic operators}

\begin{algorithm}[tbp]
\caption{Update of quantum genes states in QIGA2}
\begin{algorithmic}[1]

\FOR{$i$ in $0, \ldots, |Q|-1$}
	\FOR{$j$ in $0, \ldots, N/2$}
		\STATE 		$q'=[0~~0~~0~~0]^T$\;
		\STATE 		$bestamp \gets j\textrm{-th pair of binary genes in } b$ as decimal\;
		\STATE 		$sum \gets 0$\;
		\FOR{$amp$ in $\{0, 1, 2, 3\}$}
				\IF{$amp \ne bestamp$}
					\STATE $q'[amp] \gets \mu \cdot q_{ij}$\;
					\STATE $sum \gets sum + \left(q'[amp]\right)^2$\;
				\ENDIF
		\ENDFOR
		\STATE	$q'[bestamp] \gets \sqrt{1 - sum} $\;
		\STATE	$q_{ij} \gets q'$\;
		
	\ENDFOR
\ENDFOR

\end{algorithmic}
\end{algorithm}

The second original element of the QIGA2 algorithm is the use of genetic operators.
In the QIGA1 algorithm genetic operators are created by unitary $2\times 2$ quantum gates
(thanks to the limiting of the amplitudes to a set $\mathbb{R}$, they become just matrices of a normalised state vector rotation on a plane).
By contrast, in the QIGA2 algorithm the genetic operators can be described by $4\times 4$ quantum gates in the quantum-mechanical sense. 


The pseduocode for the operation of measuring the states of a 2-qubit quantum register $q_{ij}=\alpha_0|00\rangle+\alpha_1|01\rangle+\alpha_2|10\rangle+\alpha_3|11\rangle$
${}=[\alpha_0~~\alpha_1~~\alpha_2~~\alpha_3]^T$ coding a pair of classic binary genes is presented in the Algorithm 2.
The observation function returns strings of binary genes 00, 01, 10 and 11 with a probability of $|\alpha_0|^2$, $|\alpha_1|^2$, $|\alpha_2|^2$ oraz $|\alpha_3|^2$
respectively. 


Algorithm 3 presents the pseudocode of the proposed new genetic operator (observing the state of a 2-qubit quantum gene) in QIGA2.
Index $i$ of the main operator's loop iterates through all the individuals in the quantum population $q_0, \ldots, q_{|Q|-1}$.
Index $j$ iterates through all the consecutive \emph{pairs} of genes $j \in \{0, 1, \ldots N/2\}$ of a given quantum individual $q_i$.
Within these loops, a new state $q'$ of the quantum gene pair number $j$ of the character $q_i$ is calculated. 


The update is performed in the following manner: If the amplitude $\alpha_{amp} (amp \in \{0,1,2,3\})$ does not correspond to a $j$-th pair
of bits of the currently best found individual $b$, the amplitude is decreased (amplitude contraction) according to the rule:
$q_{ij}[amp]' = \mu \cdot q_{ij}[amp]$, where $\mu \in (0,1)$ is the algorithm's parameter. 
The amplitude of a pair of bits on position $j$ in the best individual $b$ is modified to preserve the normalization condition of the state vector (i.e. unit sum of probabilities
$\sum_{amp=0}^{3}|\alpha_{amp}|^2=1$).


Based on empirical experiments it has been established that the the best efficacy of an algorithm is achieved for the parameter value $\mu \approx 0.99$.
In order to further increase the efficacy, the value of the parameter $\mu$ in the QIGA2 algorithm might be subject to metaoptimalisation
(similarly to [19,20,21,31]).

\begin{figure}[tbp]
	\centering
	\includegraphics[width=.9\hsize]{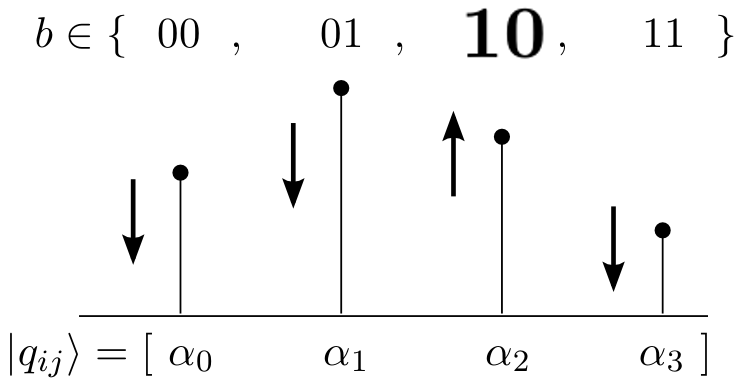}
	\caption{The new quantum genetic operator idea in QIGA2}
\end{figure}

The way the new operator works is illustrated in Figure 5. The vertical bars represent probability amplitudes $|\alpha_0|^2, |\alpha_1|^2, |\alpha_2|^2, |\alpha_3|^2$.
If on the position $j\in\{0, 1, \ldots, \frac{N}{2}\}$ of the individual $b$ there is a pair of bits 10, all the amplitudes get contracted by the factor of $\mu$,
except for $\alpha_2$ which will increase. If on the position $j$ of the individual $b$ there is a pair of bits 00, all the amplitudes get contracted by the factor of $\mu$, except for $\alpha_0$, which will increase etc.
Therefore, the only amplitude that increases is the one that corresponds to the $j$-th pair of bits in the best individual $b$. 
This makes the algorithm converge to the best individual $b$ gradually, but also doing global exploration of the search space $X$.


Simplicity is an unquestionable advantage of the QIGA2 algorithm. It is not only simpler than QIGA1, but
also less complicated than its later modified variants, whose authors also tried improve on the efficacy of the original algorithm. 
It should be noted that in QIGA2 the use of the Lookup Table (used in the original Han's QIGA1 algorithm[2]) has been eliminated completely.


\section{Numerical experiments}

For empirical comparison of the algorithms performance,
there was used a benchmark consisting of a broad set of 20 recognized combinatorial optimization problems
of different sizes
$N\in\{48, 90, \ldots, 1000\}$,
encoded in the form of the NP-complete SAT.
Objective of the combinatorial optimization process was to find
a binary string that have maximum fitness value.
The benchmark has been taken from [32], and all details about
the test functions are available there.

The compared algorithms were SGA[30], the original QIGA1[2],
the QIGA1 tuned in meta-optimization process[31]
and the authors' QIGA2.
Numerous publications to date present that QIGA1 is more effective than other modern stochastic search methods and hence
its comparison to other algorithms has been omitted in this paper as it has been assumed to be superior to other newest algorithms.

The classic SGA algorithm was run with its typical parameters values taken from [30]:
the population
size was set to 100 individuals (binary solutions), evolving for
50 generations. Thus, the total number of fitness evaluations
was equal in all algorithms, and the stopping criterion was
maximum number of fitness evaluations $MaxFE=5000$.
In SGA, single point crossover
operator with probability $P_c = 0.65$ and mutation operator
with probability $P_m = 0.05$ were used. The selection was
based on the roulette wheel method.
Implementation of SGE algorithm was taken from the external PyEvolve library [26]
The parameters for the original QIGA1 algorithm were taken from [2]
as were the parameters for the tuned QIGA1, where the only changed parameters were those that had been meta-optimized.
The QIGA2 algorithm was run with the value of the parameter $\mu=0.9918$.
For each of the test problem, each algorithm was run 50 times.  

\begin{figure}[tbp]
\includegraphics[width=\hsize]{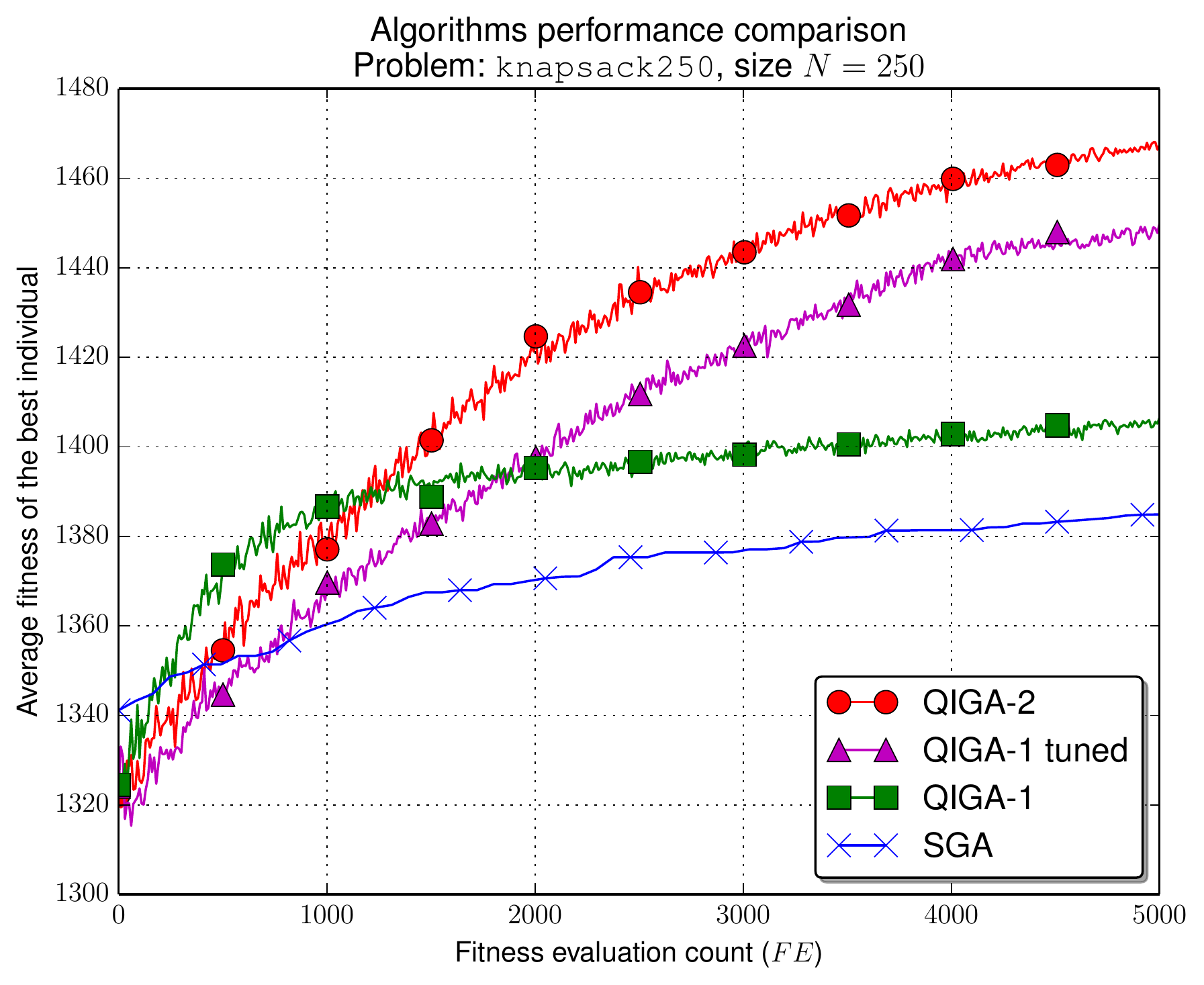}
\caption{Detailed comparison of the algorithms for a selected problem \texttt{knapsack}, size $N=250$\label{fig:knapsack250_comparison}}
\end{figure}

\begin{figure}[tbp]
\includegraphics[width=\hsize]{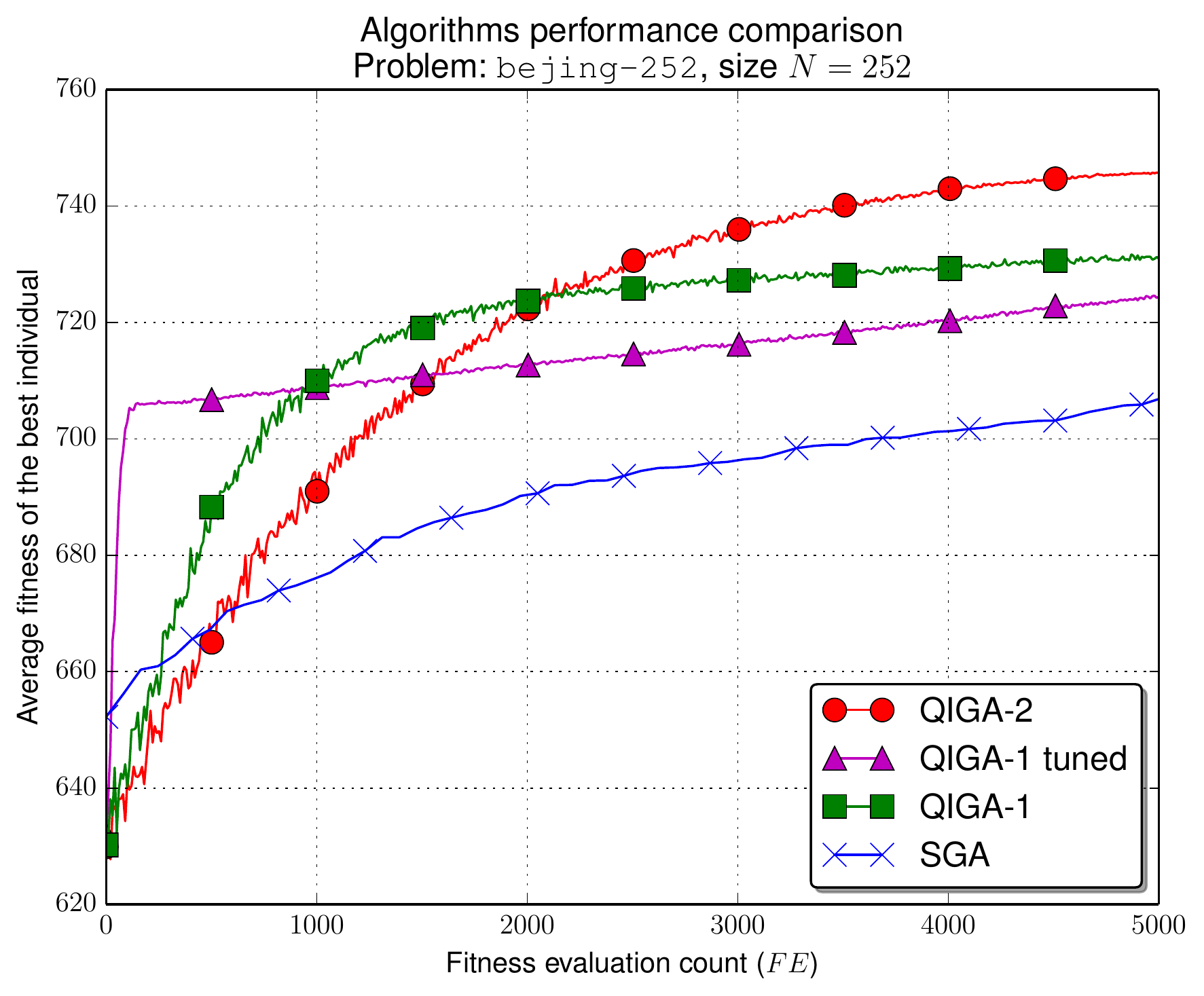}
\caption{Detailed comparison of the algorithms for a selected problem \texttt{bejing}, size $N=252$}
\end{figure}

\begin{figure}[tbp]
\includegraphics[width=\hsize]{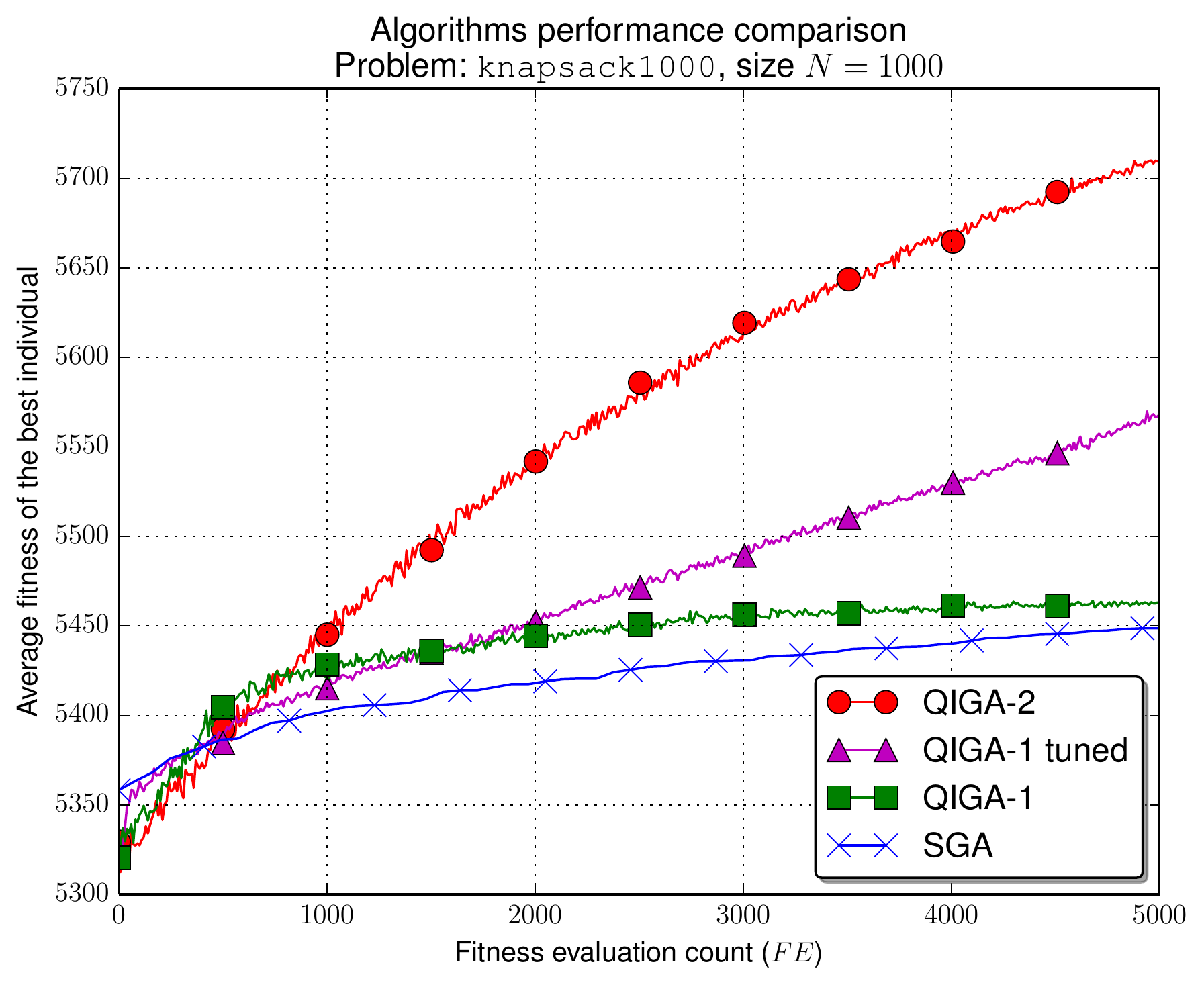}
\caption{Detailed comparison of the algorithms for a selected problem \texttt{knapsack}, size $N=1000$}
\end{figure}

As a means for evaluating the algorithms efficacy the authors used the fitness value of the best individual
after the number of generations which reached the 5000th call of the fitness evaluation function.  
Because of stochastic nature of evolutionary algorithms,
that value was later averaged for 50 runs of a given algorithm. 

In Table 1, the results for each algorithm are presented.
\textbf{In 17 out of 20 test problems (85\%), the authors' QIGA2 algorithm presented on average a better solution than both the original and the tuned QIGA1 algorithm.}
Table 2 presents a ranking of the compared algorithms ordered according to the number of test problems for which a given algorithm achieved the best result comparing to algorithms. 
Figures 6-8 present a detailed comparison of the algorithms performance for three selected test problems of size $N=250$, $N=1000$ and $N=252$. The graph shows the mean value of the best solution found by each of the algorithms versus number of calls of the individual fitness evaluation function. The presented data is averaged for 50 runs of each algorithm.



Thanks to the simplification of the algorithm and, specifically, owing to the elimination of the LookupTable,
also \textbf{the implementation of QIGA2 algorithm is 15-30\% faster than that of the QIGA1}
(the algorithms were implemented in the same programming languages, with the same compiler versions and on the same hardware platforms). 

\begin{table}[tbp]
\caption{Algorithms efficacy comparison for various problems of different size $N\in\{48, \ldots, 1000\}$\label{tab:combinatorial-final-comparison}}
\begin{tabular}{|c|>{\centering}p{7mm}|>{\centering}p{9mm}|c|c|>{\centering}p{10mm}|}
\hline

\textbf{Problem}&\textbf{Size~$N$}&\textbf{SGA}&\textbf{QIGA-1}&\textbf{QIGA-1tuned}&\textbf{QIGA-2}\tabularnewline
\hline
anomaly&48&251.4&252.55&254.65&\cellcolor{brick}{\textbf{\textcolor{white}{\textbf{255.25}}}}\tabularnewline
\hline
sat&90&284.9&289.2&293.2&\cellcolor{brick}{\textcolor{white}{\textbf{293.7}}}\tabularnewline
\hline
jnh&100&826.15&831.05&\cellcolor{brick}{\textcolor{white}{\textbf{839.05}}}&836.05\tabularnewline
\hline
knapsack&100&577.709&578.812&592.819&\cellcolor{brick}{\textcolor{white}{\textbf{596.476}}}\tabularnewline
\hline
sat&100&408.6&413.6&418.6&\cellcolor{brick}{\textcolor{white}{\textbf{419.7}}}\tabularnewline
\hline
bejing&125&297.35&302.1&305.35&\cellcolor{brick}{\textcolor{white}{\textbf{306.2}}}\tabularnewline
\hline
sat-uuf&225&886.75&898.25&\cellcolor{brick}{\textcolor{white}{\textbf{921.65}}}&921.5\tabularnewline
\hline
knapsack&250&1387.916&1406.528&1449.905&\cellcolor{brick}{\textcolor{white}{\textbf{1467.407}}}\tabularnewline
\hline
sat1&250&981.45&995.15&1021.2&\cellcolor{brick}{\textcolor{white}{\textbf{1023.1}}}\tabularnewline
\hline
sat2&250&982.95&994.6&1019.1&\cellcolor{brick}{\textcolor{white}{\textbf{1020.6}}}\tabularnewline
\hline
sat3&250&984.2&994.3&\cellcolor{brick}{\textcolor{white}{\textbf{1021.3}}}&1019.7\tabularnewline
\hline
bejing&252&709.85&731.0&724.4&\cellcolor{brick}{\textcolor{white}{\textbf{745.75}}}\tabularnewline
\hline
parity&317&1141.65&1158.2&1179.35&\cellcolor{brick}{\textcolor{white}{\textbf{1180.75}}}\tabularnewline
\hline
knapsack&400&2209.925&2222.160&2284.969&\cellcolor{brick}{\textcolor{white}{\textbf{2334.494}}}\tabularnewline
\hline
knapsack&500&2803.266&2812.740&2869.774&\cellcolor{brick}{\textcolor{white}{\textbf{2929.469}}}\tabularnewline
\hline
bejing&590&1263.8&1343.15&1284.0&\cellcolor{brick}{\textcolor{white}{\textbf{1353.2}}}\tabularnewline
\hline
lran&600&2310.9&2330.35&2386.8&\cellcolor{brick}{\textcolor{white}{\textbf{2398.95}}}\tabularnewline
\hline
bejing&708&1510.65&1605.9&1523.15&\cellcolor{brick}{\textcolor{white}{\textbf{1611.55}}}\tabularnewline
\hline
knapsack&1000&5451.656&5462.718&5568.234&\cellcolor{brick}{\textcolor{white}{\textbf{5709.116}}}\tabularnewline
\hline
lran&1000&3819.65&3848.4&3918.5&\cellcolor{brick}{\textcolor{white}{\textbf{3937.3}}}\tabularnewline

\hline
\end{tabular}

\end{table}

\begin{table}[tbp]
\caption{Ranking of the compared algorithms\label{tab:combinatorial-ranking}}
\centering
\begin{tabular}{|c|l|p{25mm}|}
\hline
\textbf{Rank} & \textbf{Algorithm} & \textbf{No. of Best\newline
Solutions}\tabularnewline
\hline
1 & QIGA2 & 17\tabularnewline
\hline
2 & QIGA-1 tuned & 3\tabularnewline
\hline
3 & QIGA-1 & 0\tabularnewline
\hline
4 & SGA & 0\tabularnewline
\hline
\end{tabular}

\end{table}

\section{Conclusions}




In this paper, fundamentals of Higher-Order Quantum-Inspired Genetic Algorithms have been presented.
The authors' original QIGA2 algorithm has been created on the basis of this theory.
The paper introduces a new way of representing solutions
using adjacent quantum registers
and a new genetic operator working in the space of a higher dimension
in quantum-mechanical sense. 
Based on empirical data gathered from 20 varied deceptive test problems of diverse sizes $N \in \{48, \ldots, 1000\}$,
it has been shown that the authors' QIGA2 algorithm achieves a better performance than both the original and the tuned QIGA1 algorithms.
Consequently, it shows that using quantum order $r=2$ is a method for improving the performance of Quantum-Inspired Genetic Algorithms.
Further investigations may include the application of the presented theory
of Higher-Order Quantum-Inspired Genetic Algorithms to a very important field of problems of numerical optimization.


%


\clearpage
\clearpage

\end{document}